%% file: iclr2023_conference.tex
\pdfoutput=1
\documentclass[11pt]{article}

\usepackage{emnlp-2023}

\usepackage{times}
\usepackage{latexsym}

\usepackage[T1]{fontenc}
\usepackage[utf8]{inputenc}

\usepackage{microtype}

\usepackage{inconsolata}

\input{math_commands.tex}

\usepackage{booktabs}
\usepackage{comment}
\usepackage{hyperref}
\usepackage{url}
\usepackage{algorithm}
\usepackage{algorithmic}
\usepackage{multirow}
\usepackage{graphicx}
\usepackage{xcolor}
\usepackage{color, colortbl}

\newcommand{\nllb}{\textsc{nllb}}
\newcommand{\holisticbias}{\textsc{HolisticBias}}
\newcommand{\alti}{\textsc{alti+}}
\newcommand{\flores}{\textsc{Flores-200}}
\usepackage{newfloat}
\usepackage{listings}
\floatstyle{ruled}
\newfloat{listing}{tb}{lst}{}
\floatname{listing}{Listing}
\begin{document}
\title{Toxicity in Multilingual Machine Translation at Scale}

% Authors must not appear in the submitted version. They should be hidden
% as long as the \iclrfinalcopy macro remains commented out below.
% Non-anonymous submissions will be rejected without review.

\author{Marta R. Costa-juss\`a$\dag$, Eric Smith$\dag$, Christophe Ropers$\dag$, Daniel Licht$\dag$, Jean Maillard $\dag$\\ 
\textbf{Javier Ferrando$^*$ \& Carlos Escolano$^*$ }\\
$\dag$Meta AI \\
$^*$Universitat Polit\`ecnica de Catalunya\\
\texttt{\{costajussa,ems,chrisropers,dlicht,jeanm\}@meta.com} \\
\texttt{\{javier.ferrando.monsonis,carlos.escolano\}@upc.edu} \\
}

% The \author macro works with any number of authors. There are two commands
% used to separate the names and addresses of multiple authors: \And and \AND.
%
% Using \And between authors leaves it to \LaTeX{} to determine where to break
% the lines. Using \AND forces a linebreak at that point. So, if \LaTeX{}
% puts 3 of 4 authors names on the first line, and the last on the second
% line, try using \AND instead of \And before the third author name.

\newcommand{\fix}{\marginpar{FIX}}
\newcommand{\new}{\marginpar{NEW}}

%\iclrfinalcopy % Uncomment for camera-ready version, but NOT for submission.
\floatstyle{ruled}

\maketitle
\begin{abstract}
Machine Translation systems can produce different types of errors, some of which are characterized as \textit{critical} or \textit{catastrophic} due to the specific negative impact that they can have on users. %Automatic or human evaluation metrics do not necessarily differentiate between such critical errors and more innocuous ones. 
In this paper we focus on one type of critical error: added toxicity. We evaluate and analyze added toxicity when translating a large evaluation dataset (\holisticbias{}, over 472k sentences, covering 13 demographic axes) from English into 164 languages. An automatic toxicity evaluation shows that added toxicity across languages varies from 0\% to 5\%. The output languages with the most added toxicity tend to be low-resource ones, and the demographic axes with the most added toxicity include sexual orientation, gender and sex, and ability. We also perform human evaluation on a subset of 8 translation directions, confirming the prevalence of true added toxicity. %Overall, toxic words in the translation are aligned to descriptor words in the source sentence 76\% of the time. % When analysing the input attributions of the toxicity, the \alti{} method shows that the source contribution average is 1.3\% lower with words adding toxicity.
%When evaluating the automatic toxicity measure on seven languages with various rates of added toxicity, human analysis reveals that this method has a false negative rate lower than 3\% and a false positive rate less than 1\% (except for two languages that are above 35\%). 
 We use a measurement of the amount of source contribution to the translation, where a low source contribution implies hallucination, to interpret what causes toxicity.  Making use of the input attributions allows us to explain toxicity, because the source contributions significantly correlate with toxicity for 84\% of languages studied.
%We observe that the source contribution is somewhat correlated with toxicity but that 45.6\% of added toxic words have a high source contribution, suggesting that much of the added toxicity may be due to mistranslations. % or semantically related translations with a wrong level of offensiveness. 
% Combining the signal of source contribution level with a measurement of translation robustness allows us to flag 22.3\% of added toxicity, suggesting that added toxicity may be related to both hallucination and the stability of translations in different contexts. 
Given our findings, our recommendations to reduce added toxicity are to curate training data to avoid mistranslations, mitigate hallucination and check unstable translations.
\end{abstract}

% \item Mitigation, flag translations {A comment about this: source contribution may or may not be correlated with toxicity in the sense that we can have a very low source contribution and a bad translation that it is not toxic... also we can have a high source contribution and a toxic word translation if this appears as such in the training data. in that sense, source contribution maybe used to filter training data if we do not want to have translations like queer to ``marica''.  one interesting thing, as future work, would be to quantify specific high source contribution benign words translated as toxic in the training data.}

{\color{olive}\textit{WARNING: this paper contains examples of toxicity that may be offensive or upsetting in nature.}}

\section{Introduction}
\label{sec:introduction}

Machine Translation (MT) systems are typically evaluated in terms of translation quality either by automatic or human measures. Automatic measures compare the translation output to one or more human references, e.g., \citet{bleu,chrf,lo-2019-yisi,rei-etal-2020-comet,sellam-etal-2020-bleurt}, or pretrained embeddings, e.g., \citet{lo-2019-yisi, yankovskaya-etal-2019-quality}. Human measures use annotators to rank translation outputs, e.g., \citet{xsts,akhbardeh-etal-2021-findings}. However, most of these evaluation strategies % based on previous two strategies
tend to lack discrimination between venial and critical errors. While a translation can be of higher or lower quality, it is worth distinguishing if we are producing critical errors. %\cite{vilar:2006} is an example of a taxonomy for translation errors in general. 
The critical error detection task aims at predicting sentence-level binary scores indicating whether or not a translation contains a critical error (not limited to toxicity) \cite{specia-etal-2021-findings}, and \citet{sharou-specia-2022-taxonomy} provide a taxonomy to classify critical errors. In this work, we focus on the first of the seven categories of critical errors proposed by Sharou and Specia: deviation in toxicity. More specifically, we evaluate cases of \textit{added toxicity}, by which we mean toxicity that is not present in the source but is introduced in the translation output. Our definition of added toxicity differs from the broader category of \textit{deviation in toxicity} in that it does not cover cases of deletion. 
%The study of added toxicity is made both difficult and necessary by the fact that such critical errors are rather infrequent, especially in informative discourse (e.g., Wikipedia, news), but have a significant impact on translation safety and user trust. 
\citet{nllb} evaluates potential added toxicity on machine translations of the \flores{} benchmark dataset using wordlist-based detectors. Such detectors are known for their limitations in over-detecting terms that are toxic only in specific contexts. Nevertheless, the overall prevalence of potential added toxicity remains low when evaluating translations of formal sentences such as those in \flores{}, which makes it difficult to draw conclusions as to this specific aspect of a model's performance. %For instance, previous work by the No Language Left Behind team \citep{nllb} contextualizes the need for low-resource language translation, expands a previous quality evaluation benchmark (FLORES) to 200 languages, many of which are considered low-resource, and evaluates translation safety by using toxicity detectors based on wordlists for all language directions. However, machine translation outputs corresponding to source sentences that pertain to informative and formal discourse (such as the FLORES-200 evaluation benchmark sentences) produce many examples of candidate added toxicity but very few examples of confirmed added toxicity. Wordlist-based detectors are indeed known for their limitations when it comes to over-detecting items that are toxic only in specific contexts. This overall low prevalence of added toxicity makes it difficult to draw conclusions as to this specific aspect of a model's performance.

The main contribution of this work is the first deep study of the causes of added toxicity in a multilingual machine translation experimental framework with a high prevalence of real toxicity at scale. For this purpose, we combine the previously defined toxicity detection methodology \citep{nllb}, the controlled evaluation dataset \holisticbias{} \citep{holisticbias}, and the \alti{} interpretability method \citep{alti+}. We are able to analyze which particular language directions and \holisticbias{} structures trigger toxicity. Moreover, we perform a human evaluation of the toxicity detection methodology for a subset of eight out-of-English translation directions, and we find that the false positive rates are below 1\% in five translation directions. % and above 35\% in two directions. 
False negatives are below 3\% in all translation directions. Finally, we demonstrate an interaction between the source contribution, the robustness of translations, and toxicity. We use \alti{} to 
% Finally, we use \alti{} to show input attributions for toxicity. We 
observe that 45.6\% of the toxic translations have a high source contribution, which hints that much of these toxic translations may be caused by mistranslations, and that the rest may be
 %or semantically related translations with a wrong level of offensiveness. 
%The rest of the toxic terms have a low source contribution, which is
correlated with hallucination \citep{alti+}. This suggests that hallucination may add toxicity. We use Gini impurity \citep{breiman1996some}, a common splitting criterion in decision trees, to measure the relative amount of diversity (i.e. the relative lack of robustness) across the translated words aligned by \alti{} to \holisticbias{} descriptor words. A combination of a low amount of source contribution and a high Gini impurity across translations corresponds to a rate of toxicity roughly twice as high as the baseline rate.
%allows us to flag toxic translations at inference time, catching 22.3\% of the toxicity insertions. 
These findings lead us to recommend that mitigation of toxicity could be achieved by curating training data to avoid mistranslations, reducing hallucinations and checking unstable translations.  
%Since a low source contribution correlates with hallucination \cite{alti+}, we can say that toxicity is somehow related to hallucination. However, in addition to hallucinations, toxicity may also be produced by other phenomena such as mistranslations or semantically related translations with the wrong level of offensiveness as we discuss in section \ref{sec:def} and \ref{sec:sourcecontribution}.

\section{Definitions and Background}
\label{sec:def}

\textbf{Definitions} %In this work, we explore one category of critical error in the translation output: deviation in toxicity. 
\citet{sharou-specia-2022-taxonomy} define deviation in toxicity as ``instances where the translation may incite hate, violence, profanity or abuse against an individual or a group (a religion, race, gender, etc.) due to incorrect translations''. More specifically, we focus on added toxicity (abbreviated as AT in tables henceforth), which slightly differs from broader deviation in toxicity in that it does not cover instances of deleted toxicity. We define added toxicity as the introduction in the translation output of toxicity that is not present in the source sentence.
%We hypothesize that added toxicity may occur in the form of mistranslation, semantically related translation and a different level of offensiveness, or hallucination. Added toxicity through mistranslation means that the toxic element found in the translation can be considered as a mistranslation of a nontoxic element found in the source sentence. An example of mistranslation can be seen in Figure \ref{fig:hallucinationmistranslation}, where the English word \textit{gangly} is mistranslated into the Catalan toxic word \textit{malparit} (meaning \textit{bastard} or \textit{fucker}). Added toxicity through a semantically related translation and a different level of offensiveness comes from the fact that the toxic element semantically corresponds to a nontoxic element found in the source sentence but it adds toxicity. An example of this case can be seen in Figure \ref{fig:hallucinationmistranslation}, where the English word \textit{queer} is translated into the Spanish toxic word \textit{marica}. Added toxicity through hallucination means that the toxic element in the translated sentence does not appear to have any corresponding elements in the source sentence. An example of hallucination can be seen in Figure \ref{fig:hallucinationmistranslation}, where the English word \textit{chubby} gets translated as \textit{grosse} (meaning \textit{fat} or \textit{big}), and the word \textit{chatte} (\textit{pussy}) appears to have no corresponding words in the source sentence. 

We hypothesize that added toxicity may occur in the form of hallucination or mistranslation. Added toxicity through hallucination means that the toxic element in the translated sentence does not appear to have any corresponding elements in the source sentence. An example of hallucination can be seen in Figure \ref{fig:hallucinationmistranslation} (Sentence 1), where the English word \textit{chubby} gets translated as \textit{grosse} (meaning \textit{fat} or \textit{big}), and the word \textit{chatte} (\textit{pussy} or \textit{pussycat}) appears to have no corresponding words in the source sentence. Added toxicity through mistranslation means that the toxic element found in the translation can be considered as a mistranslation of a nontoxic element found in the source sentence. An example of mistranslation can be seen in Figure \ref{fig:hallucinationmistranslation} (Sentence 2), where the English word \textit{gangly} is mistranslated into the Catalan toxic word \textit{malparit} (meaning \textit{bastard} or \textit{fucker}). 
\input{contributions}
\begin{comment}
\begin{table}[]
    \centering
    \begin{tabular}{lCl}
    \toprule
        Source sentence & [eng] & \textit{I don't like scrubby children.} \\
        Translation & [cat] & \textit{No m'agraden els nens malparits.} \\
        Back-translation & [eng] & \textit{I don't like bastard children.} \\
    \bottomrule
    \toprule
        Source sentence & [eng] & \textit{.} \\
        Translation & [spa] & \textit{} \\
        Back-translation & [eng] & \textit{} \\
    \bottomrule
    \toprule
        Source sentence & [eng] & \textit{I love being chubby.} \\
        Translation & [fra] & \textit{J'adore être une grosse chatte.} \\
        Back-translation & [eng] & \textit{I love being a fat pussy.} \\
    \bottomrule
    \end{tabular}
    \caption{Example of added toxicity through mistranslation (top), semantically related translation and different level of offensiveness (middle) and hallucination (bottom)}
    \label{table:mistranslation}
\end{table}
\end{comment}
%\begin{table}[]
%    \centering
%    \begin{tabular}{lCl}
%    \toprule
%        Source sentence & [eng] & \textit{I love being chubby.} \\
%        Translation & [fra] & \textit{J'adore être une grosse chatte.} \\
%        Back-translation & [eng] & \textit{I love being a fat pussy.} \\
%    \bottomrule
%    \end{tabular}
%    \caption{Example of added toxicity through hallucination}
%    \label{table:hallucination}
%\end{table}
When it comes to the level of added toxicity in translation directions, we define high-, mid-, and low-toxicity translation directions as the ones that have levels of added toxicity above 0.5\%, between 0.1\% and 0.5\%, and below 0.1\%, respectively. These percentages are computed following the approach in section \ref{sec:quant}. 
We differentiate between high- and low-resource languages following \citet{nllb}. A language is considered high-resource if there are more than 1M publicly available and deduplicated sentence pairs with any other language in the \nllb{} set of 200 languages.

\textbf{Toxicity detection methodology} \citet{nllb} propose a toxicity detection method based on wordlists for 200 languages. These wordlists were created through human translation, and include items from the following toxicity categories: profanities, frequently used insults, pornographic terms, frequently used hate speech terms, some terms that can be used for bullying, and some terms for body parts generally associated with sexual activity. 
Among their different detection methods, the authors label a sentence as toxic if it contains at least one entry from the corresponding language's toxicity word list. An entry is considered to be present in a sentence if it is either surrounded by spaces, separators (such as punctuation marks), or sentence boundaries, and thus this method would not detect words such as \textit{bass} or \textit{assistant} when looking for the toxic entry \textit{ass}. 
%As previously mentioned, wordlist-based toxicity detectors have clear limitations. However, they also have clear advantages. 
One advantage of this type of classifier is transparency, which diminishes the possibility of covering up biases \cite{xu-etal-2021-detoxifying}. Alternate methods, such as classifiers,\footnote{For instance, \url{https://www.perspectiveapi.com/}} are available for English and a few other languages but cannot be used in massively multilingual environments.

\textbf{\holisticbias{}} \holisticbias{} consists of over 472k English sentences (e.g., \textit{``I am a disabled parent.''}) used in the context of a two-person conversation. Sentences are typically created from combining a sentence template (e.g., \textit{``I am a [NOUN PHRASE].''}), a noun (e.g., \textit{parent}), and a descriptor (e.g., \textit{disabled}) from a list of nearly 600 descriptors across 13 demographic axes such as ability, race/ethnicity, or gender/sex. The descriptors can come before the noun (\textit{``I am a disabled parent.''}), after the noun (\textit{``I am a parent who is hard of hearing.''}), or in place of a separate noun (\textit{``I am disabled.''}) The noun can imply a certain gender (\textit{girl}, \textit{boy}) or avoid gender references (\textit{child}, \textit{kid}). Sentence templates allow for both singular and plural forms of the descriptor/noun phrase (\textit{``What do you think about disabled parents?''})
Other datasets consisting of slotting terms into templates were introduced by \citet{kurita2019measuring,may2019measuring,sheng2019woman,brown2020language,webster2020measuring}. The advantage of templates is that terms can be swapped in and out to measure different forms of social biases, such as stereotypical associations \citep{tan2019assessing}. Other strategies for creating bias datasets include careful handcrafting of grammars \citep{renduchintala-etal-2021-gender}, collecting prompts from the beginnings of existing text sentences \citep{dhamala2021bold}, and swapping demographic terms in existing text, either heuristically \citep{ma2021dynaboard,wang2021textflint,zhao2019gender,papakipos2022augly} or using trained neural language models \citep{qian2022perturbation}. 

\textbf{\alti{} method}\label{sec:alti} Input attributions are a type of local explanation that assigns a score to each of the input tokens, indicating how much each of the tokens contributes to the model prediction. See examples of these input attributions in Figure \ref{fig:hallucinationmistranslation}. In Neural MT, attention weights in the cross-attention module have been used to extract source-target alignments as a proxy for input attribution scores \citep{kobayashi-etal-2020-attention,Zenkel_2019,chen-etal-2020-accurate}, even though they are limited to providing layer-wise explanations. Gradient-based methods \citep{ding-etal-2019-saliency} have also been proposed: in this case the gradient of the prediction with respect to the token embeddings is computed, reflecting how sensitive a certain class is to small changes in the input. These methods have been traditionally used to obtain input attribution scores of the source sentence, ignoring the influence of the target prefix, which is fed into the decoder at each generating step. %Lately, \cite{alti+} have proposed \alti, which provides explanations for both input contexts.
\alti{} is the extension of \textsc{alti} \citep{alti} to the encoder-decoder setting in NMT. \textsc{alti} (Aggregation of Layer-wise Token-to-token Interactions) is an interpretability method for encoder-based Transformers. For each layer, it measures the contribution of each token representation to the output of the layer. Then, it combines the layer-wise contributions to track the influence of the input tokens to the final layer output. \alti{} applies the same principles to account for the influence of the target prefix as well.
For each decoding time step $t$, \alti{} provides a vector of input attributions $\vr_t \in \R^{|\sS|+|\sT|}$, where $\sS$ and $\sT$ are the input tokens of the encoder and decoder respectively. We refer to the source contribution to the prediction $t$ as the sum of the attributions of the encoder input tokens to the decoding step $t$, $\sum_{s=1}^{|\sS|} \vr_{t,s}$. The source-prediction alignment is computed by taking the input token of the encoder with highest attribution, $\argmax(\{\vr_{t,s} : s = 1,\ldots,|\sS|\})$. We exploit both source contributions and word alignments for a fine-grained analysis of toxicity as well as an approach to flag temptative toxic translations. We consider a source contribution to be low when it is smaller than a threshold of 40\%, in which case we consider the target word is much more likely to be the result of model hallucination: this threshold corresponds to a region of particularly high toxicity (section~\ref{sec:sourcecontribution}).

\section{Proposed Experimental Methodology}
\label{sec:methods}

We combine the toxicity detection methodology, \holisticbias{}, and the \alti{} method to study added toxicity in multilingual machine translation at scale.\footnote{\holisticbias{} was released under CC BY-SA 4.0 and is being used here for evaluation purposes only.} We demonstrate that \holisticbias{} is a challenging demographic dataset that triggers added toxicity in machine translation (section \ref{sec:quant}). We use a combination of the \alti{} method and the robustness of the translations to explain the causes of this toxicity (section \ref{sec:sourcecontribution}). Finally, we provide for the first time a human evaluation of the toxicity detection methodology presented in \citet{nllb} (section \ref{sec:human}). 

Following the release of highly multilingual MT models in \citet{nllb}, we are using the 3.3B dense \nllb{} model (results with the 600M distilled model are presented in Appendix \ref{apx:model}).\footnote{Models were released under CC BY-NC 4.0 and are being used here for research purposes only.} We translated the \holisticbias{} dataset, which contains 472,991 English sentences, into 164 of these 200 languages (Table~\ref{table:language_list}) in order to evaluate the toxicity of the translations. 36 languages were discarded for one of three reasons. First, for 27 languages,\footnote{Assamese, Awadhi, Bengali, Bhojpuri, Gujarati, Hindi, Chhattisgarhi, Kannada, Kashmiri, Khmer, Lao, Magahi, Maithili, Malayalam, Marathi, Meitei, Burmese, Nepali, Odia, Eastern Panjabi, Sanskrit, Santali, Shan, Sinhala, Tamil, Telugu, Thai.} tokenization on non-word characters is not sufficient to distinguish words from each another.  Even using SPM tokenization \cite{spm} on both the sentences and the toxic words list cannot provide a solution to this problem. %Poor tokenization causes overdetection because detectors get triggered by strings of characters that correspond to sub-words or syllables (an equivalent example in English would be a tokenization technique that would separate the word ``assistant'' into two tokens ``ass'' and ``istant''). In the future, poor-tokenization languages and inaccurate toxic lists should be revised.
 Second, for seven languages,\footnote{Standard Tibetan, Hungarian, Japanese, Korean, Tamasheq (Latin script), Tamasheq (Tifinagh script), Yue Chinese.} issues such as UNKs or untranslated English text prevent easy alignment of word splittings with the results of the \alti{} method. Third, for two languages,\footnote{Pangasinan and Igbo.} the toxicity lists are too inaccurate in that they include many entries whose toxicity is sensitive to context. %Basically, the lists include frequently occurring words that can be used in both toxic and nontoxic contexts. Then, we risk overdetecting toxicity by detecting them in all contexts, including those contexts where they are nontoxic. Under these circumstances, we risk ending up with plenty of false positive detections.

\section{Quantification of added toxicity}
\label{sec:quant}

In this section, we provide a coarse and fine-grained analysis of added toxicity in the experimental setting defined in previous section. %We provide a coarse and a fine-grained analysis for 164 languages.% on the demographic axes of \holisticbias{}. 
%Then, using the \alti{} method \citep{alti+}, we provide a fine-grained analysis.
%together with an analysis of the relationship between input attributions and toxicity. %Finally, we compare how the added toxicity varies in a distilled model.

\textbf{Coarse-grained analysis\label{sec:coarse_analysis}
} We use toxicity detectors to quantify toxicity per language, axis, descriptors, noun and template at the sentence level.

\begin{figure*}[h!]
\center
    \includegraphics[width=14cm]{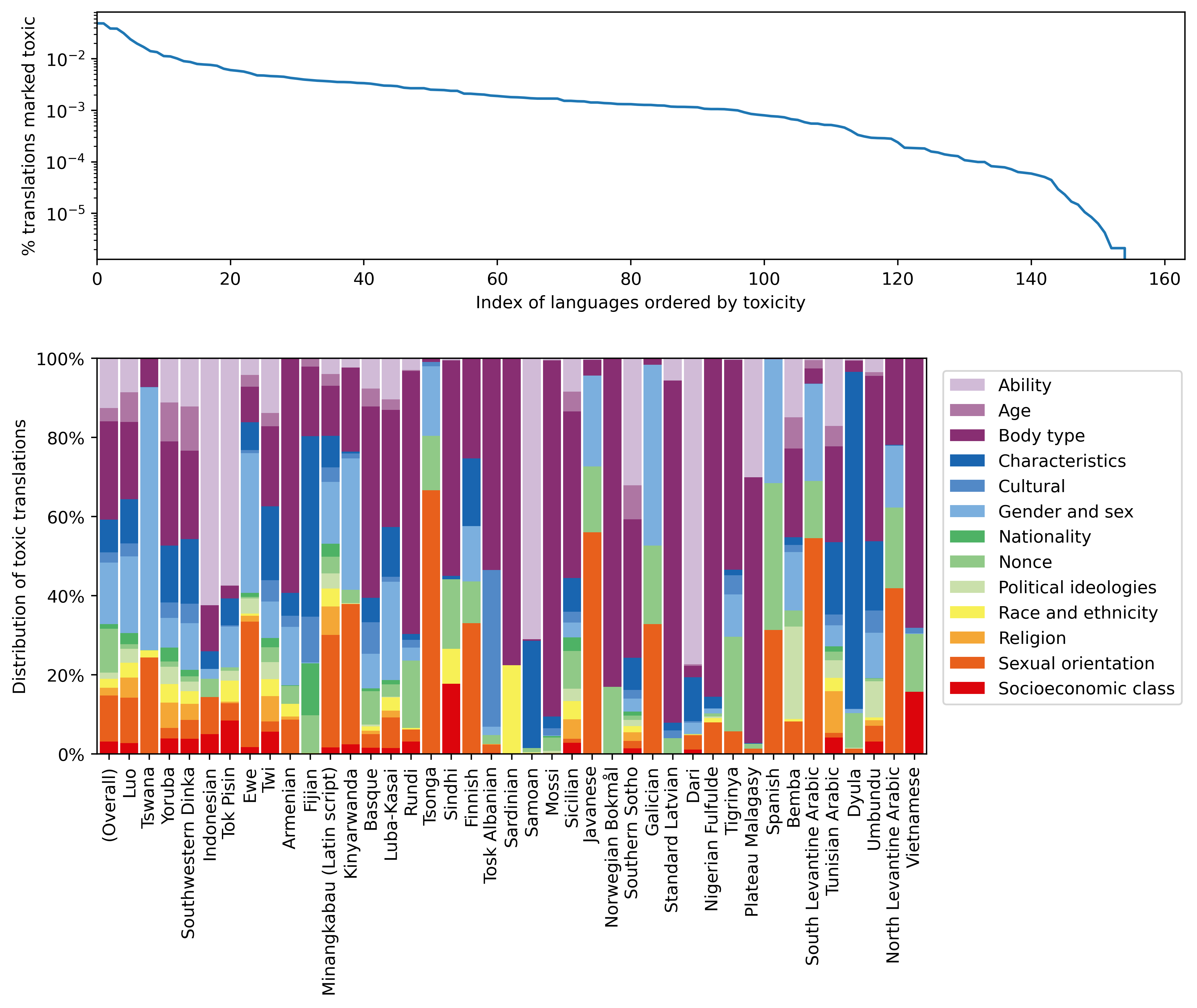}
    \caption{\textbf{Levels and types of added toxicity vary greatly as a function of language and dataset.} \textit{Top:} The fraction of translations labeled as toxic is shown as a function of language, sorted by most to least toxic, for the \flores{} and \holisticbias{} datasets. \textit{Bottom:} For \holisticbias{}, different languages have wildly different distributions of toxic terms as a function of demographic axis, with some languages' toxicity being dominated by only one or two axes. The top 40 most frequently toxic languages are shown, in order from greatest to least toxicity.}
    \label{fig:toxicity_per_lang}
\end{figure*}

\textbf{By language.} Figure \ref{fig:toxicity_per_lang} shows large variation in toxicity as a function of language and dataset. The \holisticbias{} dataset shows generally higher rates of added toxicity than \flores{}. Since we have removed any language with $>$5\% toxicity (based on malformed toxic lists), toxicity varies from 0\% to 5\%. 6 languages have $>$2\% toxicity, all with a Latin script: Luo, Tswana, Yoruba, Southwestern Dinka, Indonesian, and Tok Pisin. According to the definition of high and low resource languages in section \ref{sec:def}, all of these languages are low-resource except for Indonesian and Tswana. All but 13 languages have less than 1\% toxicity. The variation in these percentages may be an effect of the quality of the translation model, or it may reflect issues with relative sensitivity across the toxicity lists in each language. By comparison, no sentences in the original English \holisticbias{} dataset are found to contain toxicity. There is no discernible correlation between the rate of added toxicity per language and the fidelity of the translations: its Pearson's $r$ with the chrF score \citep{popovic2015chrf} is -0.06 (95\% confidence interval via bootstrapping: -0.23 to +0.12). Note that since \holisticbias{} is only available in English, we compute the quality of translations based on FLORES-200 \cite{nllb}.
% Correlation between toxicity and chrF score: /private/home/ems/GitHub/fairinternal/MultilingualBias/experiments/e2022_07_20__toxicity/00d_corr_toxicity_with_chrf.py

\textbf{By axis.} Among the 13 axes of \holisticbias{} (Figure \ref{fig:toxicity_per_lang}), the highest rates of toxicity are found in translations of terms in the nonce (non-sense) axis (3.0\% of all translations), sexual orientation (1.5\%), gender and sex (0.7\%), and ability (0.4\%). Further details are reported in Appendix \ref{apx:axisanalysis}.

\textbf{By noun.} The eight most toxic nouns all refer to parents or grandparents (\textit{parent, grandparent, father, dad, grandfather, grandpa, grandmother,} and \textit{grandma}), an effect that appears across most of the languages most likely to have toxic nouns.

\textbf{By template.} Templates expressing hate towards the demographic identity in question (\textit{``I think [PLURAL NOUN PHRASE] are the worst.''}, \textit{``I hate [PLURAL NOUN PHRASE].''}) are most likely to be rated as toxic. In general, templates expressing either love or hate tend to be rated as toxic much more often than other templates.

\textbf {Fine-grained analysis} We provide an analysis of the toxicity at the word level. \label{sec:fine_analysis}
We use \alti{} to extract the word alignment of the source sentence and the translation output. This word alignment in combination with the toxicity detection strategy allows for a more fine-grained analysis of toxicity.
Overall, in 75.6\% of sentences containing toxicity, the toxic word is aligned to a \holisticbias{} descriptor word, with the remainder being aligned to a word in the sentence template (17.4\%) or the noun (7.0\%).\footnote{We randomly select among toxic words if more than one of them is detected, as happens for 5.1\% of sentences containing toxicity.} However, this distribution varies immensely across languages (as we detail in Appendix \ref{apx:finegrained} and in Figure~\ref{fig:alignment_type_breakdown}).

\section{Phenomena causing toxicity}
\label{sec:sourcecontribution}

We explore the information from measuring the source contribution to translations, as well as the robustness in translations, in relation to toxicity.

\textbf{Input Attributions} We use the level of source contribution to confirm that toxicity can be caused by mistranslation  %a semantically related translation but a different level of offensiveness, 
and hallucination, as suggested in section \ref{sec:def}.
Note that a low source contribution is a good signal to predict hallucination \citep{alti+}, but that hallucination and toxicity are two different concepts. Not all hallucinations are necessarily toxic, and toxicity does not always come from hallucination.

\textbf{Overall contribution of the source sentence to toxicity} We use \alti{} to calculate the contribution of the source sentence to each target word in each \holisticbias{} sentence across all 164 languages. The mean source contribution, averaged across all languages, is 39.0\% for all target words, 40.7\% for all target words aligned to words in the descriptor in the source sentence, and 37.5\% for all target words identified as toxic. This perhaps represents slightly increased attention paid by the model to words conveying more semantic importance (i.e. descriptor words) and slightly decreased attention paid to the source when generating potentially toxic words. See a particular example in Figure \ref{fig:hallucinationmistranslation}: we observe that source contribution is higher in the case of a correct translation %a semantically related translation (without toxicity) 
than in the other examples where there is added toxicity. %case of mistranslation or semantically related translation with wrong level of offensiveness. All three mentioned cases have a higher source contribution than when hallucinating.

\textbf{Level of source contribution in the toxic terms} When considering the source contribution specifically to target words aligned to descriptor words in the source sentence, the mean source contribution is 40.1\% for toxic target words and 40.7\% for non-toxic target words, with 45.6\% of toxic target words and 54.8\% of non-toxic target words having a source contribution above 40\%. As mentioned in section \ref{sec:alti}, below 40\% source contribution (i.e. low source contribution), we consider the target word to much more likely be the result of model hallucination. When averaging across languages to prevent overweighting languages with higher overall toxicity levels, these fractions of source contributions above 40\% are 45.7\% for toxic target words and 54.3\% for non-toxic target words. This suggests that a good proportion of toxicity is due to mistranslations %, or to semantically related translations with a different level of offensiveness, 
in addition to hallucination. 
See examples of each of these phenomena causing toxicity and the role of source contribution in Figure \ref{fig:hallucinationmistranslation}. There, source contribution is the highest in the case of correct translation
a semantically related translation with a correct level of offensiveness; lower in the case of mistranslation; %or a semantically related translation with a different level of offensiveness; 
and lowest in the case of hallucination.
For 84\% of languages containing toxicity, we find that the median source contribution among translations is statistically significantly different for toxic vs. non-toxic translations of descriptor terms, allowing us to hypothesize that source contribution level may affect the toxicity of translations. See Appendix~\ref{apx:stats} for more details.

\textbf{Robustness of translations} 
% In order to understand which metrics correlate to increased toxicity in translations, 
We additionally compute a measure of robustness of translations to see whether that corresponds to increased toxicity as well. We compute the Gini impurity \citep{breiman1996some} (section~\ref{sec:introduction}) in the list of aligned descriptor words across the 30 nouns in the \holisticbias{} dataset, for each combination of language, descriptor, and sentence template. %The Gini impurity allows us to measure the relative lack of robustness across the translated words aligned by \alti{} to \holisticbias{} descriptor words. 
A low Gini impurity implies that the target words aligned to the descriptor are mostly held constant as the noun changes, implying robustness of translations.\footnote{Note that the Gini impurity cannot be calculated in cases where at least one of the target sentences has no words aligned to the descriptor.}
% Figure~\ref{fig:toxicity_heatmap} shows that when the aligned descriptor words are robust (low impurity), the mean source contribution is much lower to toxic aligned words than to non-toxic ones, perhaps suggesting that hallucination of toxic descriptor words may be an effect that is fairly robust to sentence perturbations such as modifications of the person noun that the descriptor refers to.

\begin{figure*}[h!]
\center
    \includegraphics[width=14cm]{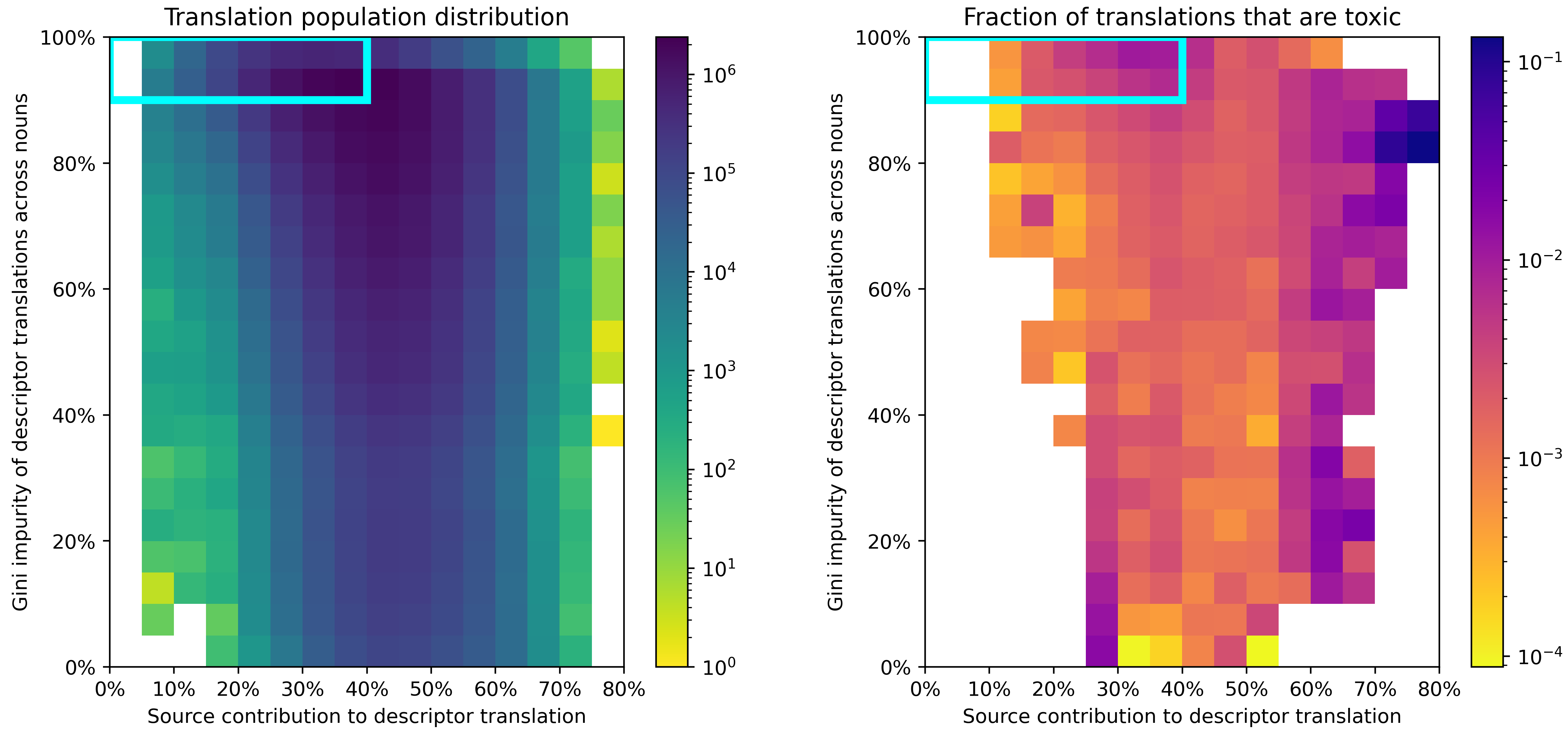}
    \caption{%\textbf{Target words aligned to demographic descriptor terms have a lower mean source contribution when they are toxic, but only if the alignment is robust across nouns.} Mean source contribution to target words aligned to descriptor words in the source sentence, as a function of whether the aligned word is flagged as toxic and the Gini impurity of the aligned descriptor words across \holisticbias{} nouns. 
    \textbf{The toxicity of descriptors in translation varies greatly as a function of both the source contribution to and the robustness of the translation.} \textit{Left:} the population distribution of the translations across all languages and \holisticbias{} sentences. \textit{Right:} the rate of toxicity of translations, with white representing no samples or 0\% toxicity. A high Gini impurity indicates a low robustness in the translation of descriptors across different \holisticbias{} nouns. Several regions have high toxicity, but many of them have few samples. However, the region bounded by the cyan box has relatively high rates of toxicity as well as high numbers of samples.}
    \label{fig:toxicity_heatmap}
\end{figure*}

% We use the source contribution to the descriptor's translation and the robustness of that translation across nouns to try to predict toxicity of descriptor words in translation. 
Figure~\ref{fig:toxicity_heatmap} shows that certain ranges of source contribution level and robustness correspond to an increased rate of toxicity. %Several of these high-toxicity regions (low source contribution and high robustness; high source contribution and low robustness) have relatively few translations, but 
Among these ranges, only the one corresponding to a low source contribution and a low level of robustness has a relatively large number of samples. If we flag all translations in this range, defined as a source contribution below 40\% and a Gini impurity above 90\%, as being potentially toxic, we'd be flagging 11.0\% of all translations but 22.3\% of all toxic translations. In this range, 0.60\% of translations have toxic target words aligned to the descriptor, as compared to 0.30\% for all translations as a whole. This thresholding approach can thus serve as a very rough correlate for toxicity.
%(Due to the relatively low overall rate of toxicity, most translations in this range are false positives: this flagging approach has a precision of 0.6\% and a recall of 22.3\%. EMS: not sure this stat matters any more now that we're not trying to claim that we're "predicting" toxicity
(Flagging translations in this range in 20 held-out languages likewise leads to 11.4\% of all translations flagged but 22.4\% of all toxic translations flagged.) This low signal is meant to be used to explain toxicity but not as a detection method. See Appendix~\ref{apx:robustness} for these results split by the level of overall toxicity in each language.

\section{Human evaluation of the toxicity detection methodology}
\label{sec:human}
%As mentioned in section~\ref{sec:introduction}, we know that the use of toxicity lists has limitations. 
Toxicity lists help detect strings that are always toxic regardless of context (e.g., \textit{fuck}, \textit{asshole}) as well as strings for which toxicity depends on context (e.g., \textit{tits}, \textit{prick}). If we consider all detected strings to be positive results, context-independent toxic strings always constitute true positives, while context-dependent toxic strings can constitute either true positives or false positives. Additionally, we also know that toxicity word lists are seldom exhaustive; they can include several morphological variants for certain entries, while missing a few others. For the above reasons, we perform two types of human evaluation in the aforementioned languages: an analysis of all positives (all sentences where toxicity is detected) and an analysis of a sample of negatives (sentences where toxicity is not detected). Language shown is in Appendix~\ref{apx:human_eval_language}.

Following our definitions in section~\ref{sec:def}, the output languages are categorized according to the prevalence of added toxicity they exhibit: high, medium, or low. We perform a manual evaluation for several languages in each category. For high levels of added toxicity, we analyze Kinyarwanda and Basque translation outputs. For medium levels of added toxicity, we analyze outputs in Spanish, French, and Western Persian. Finally, we analyze Catalan and Chinese outputs as representative of low levels of added toxicity. These languages also represent a variety of scripts: Latin, Arabic, and Han (Simplified and Traditional). 

\begin{table*}
\centering
\small
\begin{tabular}{llrrrrrrr}
%\small
\toprule
\bf{Language} & \bf{AT Level}  & \bf{Positives} & \bf{FP} & \bf{FP Rate} & \bf{Negatives} & \bf{FN} & \bf{FN Rate} \\
\midrule
Catalan & Low &  158 & 0 & 0\% & 279 & 0 & 0\% \\
Chinese (Simplified) & Low &  49 & 29 & 59.2\% & 280 & 0 & 0\% \\
Chinese (Tradidional)  & Low & 0 & 0 & n/a & 280 & 2 & 0.7\% \\
French  & Medium & 898 & 1 & 0.1\%  & 276 & 8 & 2.9\% \\
Spanish  & Medium & 1827 & 0 & 0\% & 271 & 0 & 0\% \\
Western Persian  & Medium & 1192 & 427 & 35.8\% & 273 & 0 & 0\% \\
Basque & High &  4802 & 45 & 0.9\%  & 279 & 7 & 2.5\% \\
Kinyarwanda  & High &  5264 & 313 & 5.9\%  & 255 & 0 & 0\% \\
\bottomrule
\end{tabular}
\caption{Results for the human evaluation of false positives (FP) and false negatives (FN)}
\label{table:false_positives_negatives}
\end{table*}

\textbf{Human evaluation of false positives}
The analysis of all items detected as potentially toxic (all positives) aims to separate sentences where the detected toxicity list entries are really toxic (true positives or TP) from those where context-dependent entries are used with their nontoxic meaning (false positives or FP). 
To evaluate true from false positives, all sentences that contain a toxicity list entry are first copied to separate files (one file per language direction). Each file is then shared with a linguist who is a native speaker of the translation output language. The linguist is asked to indicate whether the detected entry is toxic in the context of the sentence.
Table \ref{table:false_positives_negatives} summarizes the findings for each language. As can be seen, 5 languages have false positive rates below 1\%. Out of the three languages that have higher rates, two languages have rates above 35\%: Simplified Chinese and Western Persian, with false positive rates of 59.2\% and 35.8\%, respectively. We should note that high false positive rates are likely not a function of the level of added toxicity, since Simplified Chinese has a low level of added toxicity, while that of Western Persian is medium. 
In comparison, we report in Appendix \ref{apx:humanevalflores} the false positive analysis for the \flores{} devset. The main noticeable element presented in Table \ref{table:flores_false_positives}, beyond the high false positive rates that are observed in the \flores{} translations, is the small number of toxic entries being detected and, more particularly, the even smaller number of confirmed toxic items (4 in Kinyarwanda, 1 in Simplified Chinese, and none in the other languages). It should not be assumed that the higher rates of confirmed added toxicity found in the \holisticbias{} translations are solely due to the templated nature of the dataset, which is built by generating 780 contexts on average per descriptor. Even frequently mistranslated descriptors such as \textit{queer} (see Appendix~\ref{apx:axisanalysis}) do not produce 780 similar toxic mistranslations (374 in Kinyarwanda, 218 in French, 201 in Basque, and only 24 in Catalan).

%{\color{red} complete analysis of level of toxicity} \ccr{TO BE DISCUSSED:
%For each toxic translation, human annotators were told to mark down whether the machine translated the source demographic term to the correct axis (gender, race/ethnicity, etc.) For the average language, the mean source contribution to the target words was 52.5\% when the translation was of the correct axis and 50.6\% when it wasn't, perhaps suggesting a weak relationship between the source contribution and incorrect translations.} {redo when all issues fixed}

\textbf{Human evaluation of false negatives}
The purpose of the false negative analysis is to evaluate the likely extent to which toxicity detection may have been impeded by inconsistencies in the toxicity lists, such as missing plural or singular forms of existing entries, or missing conjugated verb forms (or any such issues related to morphological variation). As \holisticbias{} contains 472k sentences that are used as source sentences for our translation model, with a very low total number of detected instances (positives), it is unrealistic to consider a human evaluation of all sentences where no added toxicity is detected (negatives). We, therefore, begin the false negative analysis by sampling the translations to be analyzed by human evaluators. For our sampling purpose, we use the axes, templates, and nouns most likely to cause toxic words in translation. We randomly select up to 300 samples for each of the analyzed languages. 
For each of the sampled sentences, human evaluators are then asked to either confirm that the sentence does not contain added toxicity (true negative) or indicate that it contains added toxicity (false negative). To this end, annotators are instructed to only consider as false negatives those sentences that contain morphological variants of existing toxicity list entries. The goal of the false negative (FN) analysis is to ensure that the lists are comprehensive in including all derived form of the existing lemmas, which ensures the non-bias in morphological inflections compared to context-based classifiers \cite{sahoo-etal-2022-detecting}. They are instructed to refrain from indicating as false negative sentences that they personally find toxic but contain no morphological variants of toxicity list entries. 
Table \ref{table:false_positives_negatives} summarizes the results of the false negative analysis. Note that, as is the case for the false positive analysis, the FN rate for a particular language is likely not a function of its respective level of added toxicity, since French (medium AT level) has a higher false negative rate than Basque (high AT level): 2.9\% and 2.5\%, respectively. In contrast with the false positive analysis, where at least two languages show signs of substantial over-detection, the false negative analysis does not reveal such a high level of anticipated under-detection in any of the analyzed languages.

\section{Conclusions}

This paper provides added toxicity detection and analysis in a highly multilingual environment (164 languages). %For this purpose, we combine the \nllb{} toxicity detection strategy \citep{nllb}, the \holisticbias{} dataset \citep{holisticbias} and the \alti{} methodology \citep{alti+}. 
We learn that \holisticbias{} provides a good setting for analyzing toxicity because it triggers true toxicity, compared to standard previously explored datasets such as \flores{}. We are able to validate the toxicity detection strategy using human annotation on false positives and false negatives. %For a subset of 8 directions, we find that our method keeps both false positives and false negatives below 3\% except for the particular cases of false positives of Chinese and Western Persian.%, which are raised up to 59\% and 55\%, respectively. 
Additionally, we find insightful conclusions regarding the relationship between toxicity and demographic represented in \holisticbias{}, such as that the demographic axes represented in \holisticbias{} with the most added toxicity include sexual orientation, gender/sex, and ability. Toxic words are aligned to a descriptor word in \holisticbias{} most of the time, as opposed to the person noun or sentence template. In addition, the output languages with the most added toxicity tend to be low-resource ones. In the future, we want to explore if the amount of toxicity in the training data may appreciably correlate %play a bigger role in correlating 
with added toxicity. %We want to understand if the training data of a pair of languages includes a higher amount of toxicity, the model can be more robust to it (e.g., by hallucinating less) if it has a higher amount of data.
%\todo{PROBABLY DELETE SINCE WE'RE NOT LOOKING AT TOXICITY IN THE TRAINING DATA? However, we hypothesize that what creates added toxicity is unbalanced training data (i.e. toxicity is present in one side of the training but not in the other) or the amount of toxicity in the training data even if it is balanced. The latter can affect if the model tends to hallucinate. Given that \cite{nllb} filters unbalanced toxicity, the amount of balanced toxicity may play a role in correlation with hallucination.Even if the training data of a pair of languages includes a high amount of toxicity, the model can be more robust to it (e.g., by hallucinating less) if it has a higher amount of data.}
%Therefore, it is worth asking if low-resource languages tend to have more toxicity due to the fact of low training data; training data that has a higher amount of toxicity (balanced or unbalanced); or simply a combination of both factors.
Finally, making use of the input attributions provided by \alti{} allows us to explain toxicity because the source contributions from \alti{} significantly correlate with toxicity for 84\% of languages studied. We observe that 45.6\% of added toxicity has a high source contribution. Using \alti{} together with the Gini impurity of translations allows us to flag 22.3\% of toxic translations. %We find weak relationships between the contribution of the source sentence to the translation, the robustness of translations, and the toxicity, which may hint at potential causal factors between them. 
Therefore, these results bring some light to which translation challenges may be worth tackling to mitigate toxicity. The first recommendation is curating training data to avoid mistranslations that add toxicity.\footnote{{Results for some particular language pairs in Appendix \ref{apx:mitigation}}} 
This could potentially mitigate the toxicity created with high source contribution. The second recommendation is mitigating hallucinations, which may reduce toxicity in cases where we have a low source contribution. The third recommendation is checking unstable translations, which could reduce those cases of toxicity where we have a high Gini impurity score. Code and data are on GitHub.\footnote{\url{ (we are not releasing it now for anonymity reasons)}}%https://github.com/MartaCostaJussa/mtoxicity-alti-holisticbias.git}}.

%\subsubsection*{Author Contributions}
%\todo{If you'd like to, you may include  a section for author contributions as is done in many journals. This is optional and at the discretion of the authors.}

%\subsubsection*{Acknowledgments}
%The authors want to specially thank the human evaluators for their invaluable work: Aitor Ormazabal, André Niyongabo, Elahe Kalbassi, Gabriel Mejia Gonzalez, Prangthip Hansanti, Janice Lam. Also, authors want to thank Philipp Koehn for providing feedback on the final manuscript.
%\todo{Use unnumbered third level headings for the acknowledgments. All acknowledgments, including those to funding agencies, go at the end of the paper.}

\section{Limitations}

Word-based detectors are known for their limitations when it comes to over-detecting terms that are toxic only in specific contexts. Also these type of detectors have limitations in languages that do not use spaces to separate words.

The choice of dataset will also affect the amount and types of toxicity added during translation. \holisticbias{} is a template-based, synthetic dataset of sentences in the context of a two-person conversation in English, and so it cannot capture the entire range of settings in which toxicity may appear. Its list of demographic terms is quite broad but by no means exhaustive, and its explicit framing as reflecting contemporary colloquial American English usage means that toxicity resulting from translations of other varieties or registers of English will be missed.

Finally, the analysis of false negatives presented in the paper is limited to toxicity list items that may not have been detected due to morphological variation (e.g., spelling variants or missing derived word forms) because we understand that string-matching methods are particularly sensitive to such variation. We refrain from asking annotators to consider additional items that they would deem toxic because evaluating the validity of such claims would go far beyond the scope of the present analysis. 

\section{Ethics statement}
\label{sec:eth}
We follow similar ethical considerations to those stated in Subsection 7.3.5 of \citet{nllb}, and acknowledge more specifically three main areas: unintended use, biases, and safety. 
\textbf{Unintended use} Our aim is to develop techniques and metrics for the automatic detection of added toxicity in outputs of machine translation systems. We define \textit{added toxicity} in the introduction of this paper as ``toxicity that is not present in the source but is introduced in the translation output.'' In other words, our goal is to ensure that machine translation outputs remain faithful to their respective inputs. Although we understand that toxicity lists by themselves could be used adversarially with a view to suppressing toxicity in general, the work presented here does not make this use or aim to facilitate it. Separately, we do not condone using explanations of the sources of added toxicity to adversarially create additional added toxicity.

\textbf{Biases} As it is arduous to define the notion of toxicity objectively, the use of any toxicity detection method is likely to introduce biases. In the case of wordlist- and template-based methods, biases can be introduced through omissions, inconsistencies or ambiguities caused by homographs or polysemous terms. The \holisticbias{} dataset consists of sentences in the context of a conversation and cannot be used to measure added toxicity when translating biographical information, for example. Additionally, in a massively multilingual setting where most lists are built by translating the contents of an initial English-language list, biases can be due to translation errors, English-centric elements, or insufficiently diverse cultural backgrounds among available translators. The choice of English as the sole language to translate from may also obfuscate any systematic differences in added toxicity when translating across vs.\ within language families, from a language with vs.\ without strong gender markings, etc.

\textbf{Safety}
Given that the main topic of the present paper is the detection of toxicity, it would have been impossible to provide examples without including some of the detected toxic items. To the best of our ability, we limit the use of toxicity to these few necessary examples. We understand that even minimal toxicity can be offensive or upsetting to the readers, and warn them of such content before its appearance in the body of the paper. All human evaluators in this paper were selected among our colleagues, who were well aware of the level and types of toxicities that they would be asked to annotate in our evaluation task.

\bibliography{iclr,anthology,custom,iclr2023_conference}
\bibliographystyle{acl_natbib}

\appendix

\section{Appendix: Model comparison}
\label{apx:model}

%\subsection{Model comparison}

Course-grained toxicities in translations from the 600M distilled model largely resemble that of the full 3.3B dense \nllb{} model (section~\ref{sec:quant}). Some highlights:
\begin{itemize}
    \item Most of the highest-toxicity descriptors (\textit{barrel-chested, lesbian, gay, queer}) are the same between models. However, \textit{doff}, which is spelled similarly to the South African slang term \textit{dof} meaning \textit{stupid} \citep{bailey1985south}, %(Table~\todo{ref to Urban Dictionary + Merriam-Webster table}), 
    has a singularly high toxicity in the larger model but not the smaller model, perhaps reflecting a greater capacity of the larger model to intuit plausible translations of slang given misspellings.
    \item The trend of the most toxic nouns being \textit{parent/grandparent} terms is present but weaker in the smaller model: \textit{dad, father,} and \textit{parent} are the most toxic at 0.7\%, 0.6\%, and 0.5\%, respectively, followed by \textit{guy, mother, husband,} and \textit{person} at 0.4\%.
    \item \textit{``I think [PLURAL NOUN PHRASE] are the worst.''}, \textit{``I hate [PLURAL NOUN PHRASE].''}, and \textit{``I think [PLURAL NOUN PHRASE] are the best.''} are still the three most toxic sentence templates.
    \item The mean toxicity across all languages is similar, 0.38\% toxic for the small model vs.~0.36\% for the large model.
\end{itemize}

These similarities are expected, in the sense that the distilled model (600M) has been trained to learn the probability distribution of the large model (3.3B).

\section{Appendix: Axis Analysis}
\label{apx:axisanalysis}

In this appendix we report further details on the toxicity found per demographic axis from section \ref{sec:quant}.

\begin{itemize}

    \item The highest rate of toxicity is found in translations of the nonsense terms in the ``nonce'' axis: \textit{blicket, stad, mell, coodle, doff, tannin, fitch,} and \textit{tulver} \citep{soja1991ontological}. We note that some of these terms exist in English slang, and some also have toxic meanings in Merriam-Webster\footnote{\url{https://www.merriam-webster.com/}} and Urban Dictionary,\footnote{\url{https://www.urbandictionary.com/}} as well in the Corpus of Contemporary American English (COCA).\footnote{\url{https://www.english-corpora.org/coca/}}
    
    \item Sexual orientation and gender/sex terms have the next highest rates of toxicity: descriptors like \textit{queer, gay,} and \textit{lesbian} are very frequently translated as toxic, as are terms that end with the suffix \textit{``-sexual''}. In certain instances, the translation is semantically correlated to the original word, but has a much different level of toxicity than the original (for instance, translating \textit{queer} to \textit{marica} in Spanish or Catalan).
    
    \item The most commonly toxic ability terms are typically either very general, like \textit{handicapped}\footnote{The \holisticbias{} descriptor list contains terms that are often viewed as dispreferred or polarizing by members of the communities in question, and they are included to reflect the fact that these terms may still exist in models' training or evaluation data.} or \textit{disabled}, or include the words \textit{disability}, \textit{injury}, or \textit{impaired} (\textit{``with a cognitive disability''}, etc.).
    
    \item The most commonly toxic body type term is \textit{barrel-chested}, and hair terms (\textit{dirty-blonde, dark-haired}, etc.) are also often quite toxic.
    
    \item Highly toxic socioeconomic terms are \textit{trailer trash} and ones that connote poverty (\textit{broke, poor}).
    
    \item \textit{Black} is often marked as toxic, perhaps reflecting troubling and potentially racist color associations in translation. Other highly toxic terms are national-origin terms such as \textit{foreign-born, US-born,} and \textit{American-born} (perhaps indicating xenophobic translations), and often-stigmatized conditions like \textit{``an alcoholic''}, \textit{``with a gambling problem''}, and \textit{``with dementia''}.

\end{itemize}

\section{Appendix: Fine-grained analysis: variation across languages}
\label{apx:finegrained}

\begin{table*}
\centering
\small
\begin{tabular}{p{15cm}}
%\small
\toprule
Acehnese (Latin script), Afrikaans, Akan, Amharic, Armenian, Asturian, Ayacucho Quechua, Balinese, Bambara, Banjar (Arabic script), Banjar (Latin script), Bashkir, Basque, Belarusian, Bemba, Bosnian, Buginese, Bulgarian, Catalan, Cebuano, Central Atlas Tamazight, Central Aymara, Central Kanuri (Arabic script), Central Kanuri (Latin script), Central Kurdish, Chinese (Simplified), Chinese (Traditional), Chokwe, Crimean Tatar, Croatian, Czech, Danish, Dari, Dutch, Dyula, Dzongkha, Eastern Yiddish, Egyptian Arabic, Esperanto, Estonian, Ewe, Faroese, Fijian, Finnish, Fon, French, Friulian, Galician, Ganda, Georgian, German, Greek, Guarani, Haitian Creole, Halh Mongolian, Hausa, Hebrew, Icelandic, Ilocano, Indonesian, Irish, Italian, Javanese, Jingpho, Kabiyè, Kabuverdianu, Kabyle, Kamba, Kashmiri (Arabic script), Kazakh, Kikongo, Kikuyu, Kimbundu, Kinyarwanda, Kyrgyz, Latgalian, Ligurian, Limburgish, Lingala, Lithuanian, Lombard, Luba-Kasai, Luo, Luxembourgish, Macedonian, Maltese, Maori, Mesopotamian Arabic, Minangkabau (Latin script), Mizo, Modern Standard Arabic, Moroccan Arabic, Mossi, Najdi Arabic, Nigerian Fulfulde, North Azerbaijani, North Levantine Arabic, Northern Kurdish, Northern Sotho, Northern Uzbek, Norwegian Bokmål, Norwegian Nynorsk, Nuer, Nyanja, Occitan, Papiamento, Plateau Malagasy, Polish, Portuguese, Romanian, Rundi, Russian, Samoan, Sango, Sardinian, Scottish Gaelic, Serbian, Shona, Sicilian, Silesian, Sindhi, Slovak, Slovenian, Somali, South Azerbaijani, South Levantine Arabic, Southern Pashto, Southern Sotho, Southwestern Dinka, Spanish, Standard Latvian, Standard Malay, Sundanese, Swahili, Swati, Swedish, Tagalog, Tajik, Tatar, Ta’izzi-Adeni Arabic, Tigrinya, Tok Pisin, Tosk Albanian, Tsonga, Tswana, Tumbuka, Tunisian Arabic, Turkish, Turkmen, Twi, Ukrainian, Umbundu, Urdu, Uyghur, Venetian, Vietnamese, Waray, Welsh, West Central Oromo, Western Persian, Wolof, Xhosa, Yoruba, Zulu \\
\bottomrule
\end{tabular}
\caption{The 164 languages analyzed in this work, subselected from the 200 \nllb{} languages (section~\ref{sec:methods}).}
\label{table:language_list}
\end{table*}

\begin{figure*}[h!]
\center
    \includegraphics[width=14cm]{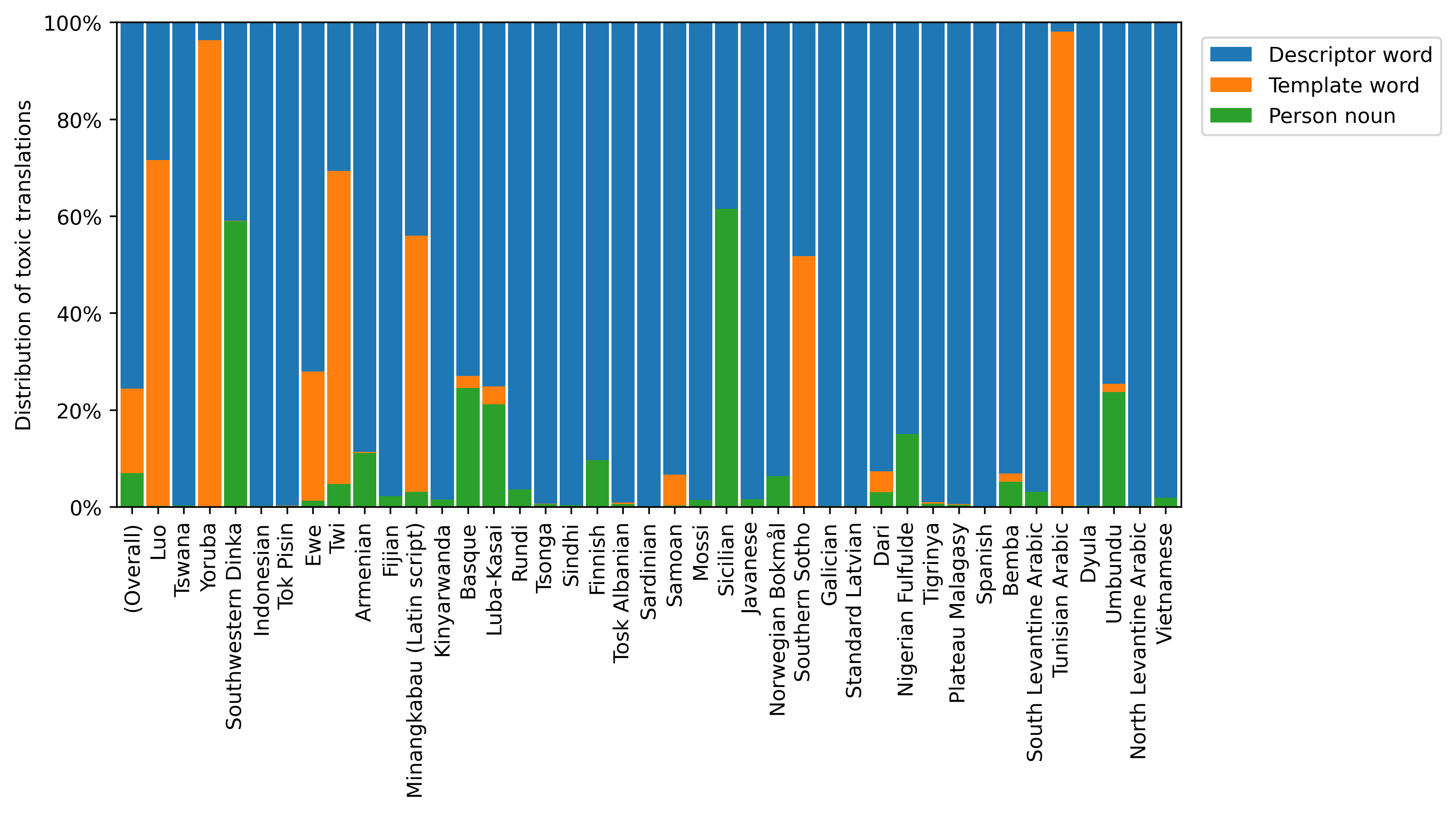}
    \caption{Distribution of target sentences found to contain toxic terms, split by the type of word in the source \holisticbias{} sentence that the toxic term is aligned to: a word in the descriptor, a word in the sentence template, or the person noun (e.g., \textit{grandma}, \textit{kid}). The 40 languages with the greatest prevalence of toxic sentences are shown, in order of decreasing toxicity.}
    \label{fig:alignment_type_breakdown}
\end{figure*}

In this section we extend the details on the fine-grained analysis from section~\ref{sec:fine_analysis} and its variation across the 164 languages analyzed (Table~\ref{table:language_list}).
\begin{itemize}

    \item \textbf{Variation in alignment types:} For instance, for Tunisian Arabic, Yoruba, Luo, Twi, Minangkabau (Latin script), and Southern Sotho, the majority of all toxic words are aligned to template words, not descriptor words (Figure~\ref{fig:alignment_type_breakdown}). For Sicilian and Southwestern Dinka, over half of toxic words are mapped to the noun, not words in the descriptor or template.%, and for Southern Sotho, 3\% of toxic words are actually mapped to the EOS token in the source, likely indicating model hallucination of these words \cite{alti+}.} % OLD: For Dholuo and Tunisian Arabic, the large majority of all toxic words are aligned to template words, not descriptor words. For Assamesse, 98\% of all toxic words are aligned to the noun, not words in the descriptor or template!
    
    \item \textbf{Template words:} 73\% of toxic words aligned to template words are aligned to \textit{worst}, followed by \textit{think} (as in \textit{``I think [PLURAL NOUN PHRASE] are the worst.''}) with 11\% and \textit{hate}, with 6\%.  However, %as with the noun distribution,
    this effect is due in large part to patterns in the alignment of toxic words in individual languages: in the cases where toxic words align to template words in the source, Yoruba and Luo almost always align to \textit{worst}, Twi to \textit{think}, and Minangkabau (Latin script) to \textit{hate}. % OLD: nearly all toxic Dholuo words aligned to template words are aligned to ``worst'', but 100\% of toxic Tunisian Arabic words aligned to template words are aligned to ``best''. For the other languages, some likewise have their aligned template word distribution dominated by one word (e.g., Armenian → ``I’m''), but some align to a greater variety of template words.

    \item \textbf{Nouns:} The 14 most common nouns that toxic words are aligned to refer to parents/grandparents: \textit{grandparents, parents, grandfathers, dads, grandpas, father, grandmothers, grandparent, dad, fathers, grandmother, grandma, grandmas,} and \textit{moms}. However, this varies by language, with Armenian having its toxic words most commonly aligned to \textit{bro, guy, individual, man, sibling,} and \textit{brother} (in 72\% of all cases of alignment to nouns). % OLD: By far, the most common nouns that toxic words are aligned to are ``sister(s)'' and ``sibling(s)''. However, this trend is entirely due to Assamese, which usually has its toxic words aligned to one of these two nouns. Other languages have different distributions of nouns that their toxic words are aligned to.
    
\end{itemize}

\section{Appendix: Statistical testing of source contribution level and toxicity}
\label{apx:stats}

% OLD, BEFORE AVERAGING ACROSS LANGUAGES: When considering the source contribution to target words specifically aligned to descriptor words in the source sentence, the mean source contribution for non-toxic target words specifically is 40.6\% (reflecting the fact that most descriptor words are not flagged as toxic), with a median of 41.0\%, and 87.2\% above a source contribution of 30\%. (As a rule-of-thumb below 30\% source contribution, the target word is much more likely to be the result of model hallucination \cite{alti+}.) For toxic words aligned to source descriptor words, the mean source contribution is 40.1\%, the median is 39.2\%, and 86.9\% are above a source contribution of 30\%.

% Given that we have produced 75M translations in which at least one target word is aligned to a \holisticbias{} descriptor word across all 171 languages, t
For each language containing toxicity, we perform a statistical test of %two statistical tests to understand the relationship between source contribution level and toxicity:
%\begin{enumerate}
    %\item 
    whether the median source contribution among all translations is the same for toxic and for non-toxic translations of descriptor terms: in 84\% such cases (i.e. for 84\% of languages tested), the null hypothesis of equal medians in Mood's median test \citep{mood1950introduction} is rejected at $p<0.05$. %If source contribution and toxicity were completely uncorrelated, we would expect to find a result at least this significant for only roughly 5\% of languages.
    %-value less than the lowest positive non-zero value of a NumPy \texttt{float64} (5e-324).
    %\item 
    We also computed whether the rate of hallucination (source contribution $<40\%$) is the same for toxic and for non-toxic translations: we use the one-sided two-proportions $z$-test to find that the null hypothesis that the rate of hallucination is equal or lower for toxic translations is rejected at $p<0.05$ for 59\% of languages that contain toxicity.
%\end{enumerate}
    These results lead us to hypothesize that the level of source contribution, and the hallucination of the model indicated by low source contribution, may play some small role in creating toxic translations. %(Or likewise, some other factor may be responsible for the observed correlations between source contribution level and toxicity.)%, but further experiments may be needed to test this hypothesis fully.
% {As section \ref{sec:coarse_analysis} shows when analysing toxicity per language, the output languages with the most added toxicity tend to be low-resource ones. However, we can discuss if what creates added toxicity is unbalanced training data (i.e. toxicity is present in one side of the training but not in the other) or the amount of toxicity in the training data even if it is balanced. The latter can affect if the model tends to hallucinate. Given that \cite{nllb} filters unbalanced toxicity, the amount of balanced toxicity may play a role in correlation with hallucination. %Therefore, it is worth asking if low-resource languages tend to have more toxicity due to the fact of low training data; training data that has a higher amount of toxicity (balanced or unbalanced); or simply a combination of both factors.} 
% {Even if the training data of a pair of languages includes a high amount of toxicity, the model can be more robust to it (e.g., by hallucinating less) if it has a higher amount of data.}
Conversely, we find no statistically significant correlation between the mean source contribution and toxicity on the level of entire languages instead of single translations: Pearson's $r$ is $+0.02$ with a 95\% confidence interval from bootstrapping of $-0.12$ to $+0.18$, and Spearman's rank correlation coefficient is $+0.13$ with a 95\% confidence interval of $-0.03$ to $+0.27$.

% (SKIPPING DUE TO ISSUES WITH THE DATA) {In addition, and in collaboration with daniel, can we do the correlation with the amount of toxicity in training? Daniel should have the data of amount of toxicity per language, it is in relation with table 29 and 30 in the paper. If we have this data, bootstrap with replacement to get confidence intervals}

\section{Appendix: Robustness of translations}
\label{apx:robustness}

%\emsc{(THIS SECTION IS ENTIRELY NEW)}

\begin{table*}[h!]
\centering
\small
\begin{tabular}{lrrrr}
%\small
\toprule
Toxicity range & Num. languages & \% toxic in region & \% toxic overall & Ratio \\
\midrule
Low ($<0.1\%$) & 57 & 0.03\% & 0.02\% & 1.25 \\
Medium ($0.1\%$ to $0.5\%$) & 68 & 0.35\% & 0.23\% & 1.50 \\
High ($>0.5\%$) & 19 & 2.42\% & 1.33\% & 1.82 \\
% \quad\textit{Very high} ($>1.0\%$) \\
\bottomrule
\end{tabular}
\caption{Amount of toxicity in the highlighted region of Figure~\ref{fig:toxicity_heatmap} as a function of the overall toxicity of each language.}
\label{table:toxicity_range_by_language}
\end{table*}

Table~\ref{table:toxicity_range_by_language} shows the amount of toxicity in the region of low source contribution and low robustness (section~\ref{sec:sourcecontribution}) split by languages that have a low, medium, or high rate of toxicity overall, given the thresholds defined in section~\ref{sec:def}. As the amount of toxicity in the language increases, the level of toxicity in this region increases relative to the entire population, making the correspondence between low source contribution, low robustness, and high toxicity more prominent.

\section{Appendix: Language shown during human evaluations}
\label{apx:human_eval_language}
The following language was shown for the human evaluation of \textbf{false positives}: \textit{``In the list of Positives (sentences where at least one toxicity list entry was detected), check the corresponding box if: (1) the detected entry is a context-independent toxic item (for example, an entry such as `fucking' in English, which is always toxic regardless of context), or (2) the detected entry is a context-dependent toxic item that is truly toxic in the context of the sentence. Don't check the corresponding box if the detected entry is a context-dependent toxic item that isn't toxic in the context of the sentence (for example, the word `prick' in the sentence `You should feel nothing but a small prick.')''}

The following language was shown for the human evaluation of \textbf{false negatives}: \textit{``In the list of Negatives (sentences where no toxicity list entries were detected), check the corresponding box if you identify a word that: (1) is toxic AND (2) is a morphological variant of a toxicity list entry. Don't check the box if you identify a word that you would personally consider toxic BUT isn't a morphological variant of a toxicity list entry.''}

\section{Appendix: Human evaluation on \flores{} dataset translations}
\label{apx:humanevalflores}
Table \ref{table:flores_false_positives} summarizes the human evaluation findings on translations of the \flores{} devtest set produced by the same model as the translations of the \holisticbias{} dataset analyzed in this paper (see section~\ref{sec:human}). As can be seen, the \flores{} devtest set produces no confirmed toxicity in six of the eight analyzed languages (the only detected entries in those languages are false positives), only 1 example of confirmed toxicity in Simplified Chinese, and 4 in Kinyarwanda. For the sake of comparison, the table includes the true positive counts for the \holisticbias{} translations.

\begin{table*}
\centering
\small
\begin{tabular}{llrrr|r}
%\small
\toprule
\bf{Language} & \bf{Positives} & \bf{FP} & \bf{FP Rate} & \bf{TP} & \bf{\holisticbias{} TP}\\
\midrule
Catalan & 1 & 1 & 100.0\% & 0 & 158 \\
Chinese (Simplified) & 2 & 1 & 50.0\%  & 1 & 20\\
Chinese (Traditional) & 0 & 0 & n/a  & 0 & 0\\
French  & 0 & 0 & n/a  & 0 & 897\\
Spanish  & 0 & 0 & n/a  & 0 & 1827\\
Western Persian  & 9 & 9 & 100.0\%  & 0 & 765\\
Basque & 2 & 2 & 100.0\%  & 0 & 4757\\
Kinyarwanda  & 23 & 19 & 82.6\%  & 4 & 4951\\
\bottomrule
\end{tabular}
\caption{Results for the human evaluation of false positives (FP) and true positives (TP) in the \flores{} dataset translations (as well as the TP count for \holisticbias{} translations in comparison).}
\label{table:flores_false_positives}
\end{table*}
    
\section{Appendix: Toxicity Mitigation}
\label{apx:mitigation}

Following our first recommendation, which is curating training data sets, we provide some initial experiments on filtering unbalanced toxicity for the 8 language pairs selected in previous sections, i.e. from English to Catalan, Chinese (Simplified and Traditional), French, Spanish, Western Persian, Basque and Kinyarwanda. For each of these pairs, we train bilingual systems with 4 different versions of the training data:
\begin{itemize}
    \item \texttt{baseline}: no toxicity filtering is performed.
    \item \texttt{max\_add\_1}: sentence pairs with added toxicity greater than 1 ($|\mathtt{src\_tox}-\mathtt{tgt\_tox}|>1$) are filtered out.
    \item \texttt{no\_add}: sentence pairs with added toxicity ($|\mathtt{src\_tox}-\mathtt{tgt\_tox}|>0$) are filtered out.
    \item \texttt{no\_tox}: a draconian baseline, used for reference purposes only, where sentence pairs with any toxicity at all ($\mathtt{src\_tox}+\mathtt{tgt\_tox}>0$) are filtered out.
\end{itemize}

The training datasets are filtered with the \texttt{stopes} library \citep{andrews-etal-2022-stopes} and tokenized using the same \texttt{sentencepiece} model \citep{kudo-richardson-2018-sentencepiece} as \citet{nllb}. The models use a transformer architecture using 6 encoder and decoder layers, 4 attention heads, embeddings of size 512 and a dropout rate of 0.3. They are trained with \texttt{fairseq} \citep{ott2019fairseq} using the Adam optimizer \citep{adam} with an inverse square root learning rate schedule with warmup, and an effective batch size of $2^{17}$ tokens. Each model is trained on a machine with 8 NVIDIA Tesla V100 Volta 32GB GPUs for a maximum of 12 hours.

The results, reported in Table \ref{table:mitigation}, display a clear trend of reduction in toxicity as filter strength is increased. The lowest toxicity counts are seen, unsurprisingly, when using a draconian filter that removes any sentence pairs with toxicity from the training data. The more reasonable approach that removes only added toxicity, \texttt{no\_add}, still manages to reduce the vast majority of detected toxicity across all inspected languages.
Table \ref{table:mitigation} includes results in chrf with Flores-200. We do not see a big difference in translation quality because of the filtering.%, except the fact that filtering tends to slightly improve translation quality. 
%We left for future work looking at how these filters affect the quality of translation.

\begin{comment}
    
\begin{table}
\centering
\small
\resizebox{\linewidth}{!}{\begin{tabular}{lrrrr}
\toprule 
\textbf{Language} & \texttt{baseline} & \texttt{max\_add\_1} & \texttt{no\_add} & \texttt{no\_tox} \\
\midrule
Basque & 1167 & 1373 & 129 & 14 \\
Catalan & 1910 & 1483 & 360 & 0 \\
Chinese (Simplified) & 101 & 14 & 1 & 0 \\
Chinese (Traditional) & 17 & 22 & 0 & 0 \\
French & 7168 & 6071 & 966 & 1182 \\
Kinyarwanda & 3697 & 3029 & 70 & 55 \\
Spanish & 2320 & 1171 & 247 & 0 \\
Western Persian & 700 & 844 & 80 & 53 \\
\bottomrule
\end{tabular}}
\caption{Toxicity detection for different bilingual systems and filters.}
\label{table:mitigation}
\end{table}

\end{comment}

\begin{table*}[h!]
    \centering
\small
\resizebox{\linewidth}{!}{\begin{tabular}{lrrrrrrrr}
\toprule 
\textbf{Language} & \multicolumn{2}{r}{\texttt{baseline}} & \multicolumn{2}{r}{\texttt{max\_add\_1}} & \multicolumn{2}{r}{\texttt{no\_add}} & \multicolumn{2}{r}{\texttt{no\_tox}} \\
\midrule
 & ETOX & chrf &  ETOX & chrf & ETOX & chrf & ETOX & chrf\\
\midrule
Basque & 1167 & \bf{50.7} & 1373 & 50.6 & 129 & 50.2 & \bf{14} & 50.4 \\
Catalan & 1910 & \bf{62.5} & 1483 &62.4 & 360 & 62.4 & \bf{0} & \bf{62.5}\\
Chinese (Simplified) & 101 & \bf{17.4} & 14 & 16.7 & 1 & 17.1& \bf{0} &17.1\\
Chinese (Traditional) & 17 & 10.9 & 22& \bf{11.3} & 0& 11.1 & \bf{0} &11.3 \\
French & 7168 & 46.4 & 6071 & \bf{63.2} & \bf{966} & 62.9 & 1182 & \bf{63.2}\\
Kinyarwanda & 3697  & 47.4& 3029& \bf{47.6} & 70 & 47.0 & \bf{55} & 47.4\\
Spanish & 2320 & \bf{50.1}& 1171 &49.3 & 247 & 50.0 & \bf{0} &49.9\\
Western Persian & 700 & 48.2 & 844 & 48.2 & 80 & \bf{48.3} & \bf{53} & 48.2\\
\bottomrule
\end{tabular}}
\caption{Toxicity detection and translation quality in terms of chrf for different bilingual systems and filters.}
\label{table:mitigation}
\end{table*}

\begin{comment}

\begin{table*}
\centering
\small
\begin{tabular}{lrrrrrrrr}
% \small
\toprule 
 & Catalan & Basque & W. Persian & French & Kinyarwanda & Spanish & Chinese (S.) & Chinese (T.)  \\
 \midrule
\texttt{baseline} & 1910 & 1167 & 700 & 7168 & 3697 & 2320 & 101 & 17 \\
\texttt{max\_add\_1} & 1483 & 1373 & 844 & 6071 & 3029 & 1171 & 14 & 22 \\
\texttt{no\_add} & 360 & 129 & 80 & 966 & 70 & 247 & 1 & 0 \\
\texttt{no\_tox} & 0 & 14 & 53 & 1182 & 55 & 0 & 0 & 0 \\
\bottomrule
\end{tabular}
\caption{Toxicity detection for different bilingual systems and filters.}
\label{table:mitigation}
\end{table*}

\end{comment}

\end{document}

%% file: math_commands.tex
\usepackage{amsmath,amsfonts,bm}

\def\eqref#1{equation~\ref{#1}}
\def\1{\bm{1}}

\def\vr{{\bm{r}}}

\DeclareMathAlphabet{\mathsfit}{\encodingdefault}{\sfdefault}{m}{sl}
\SetMathAlphabet{\mathsfit}{bold}{\encodingdefault}{\sfdefault}{bx}{n}

\def\sS{{\mathbb{S}}}
\def\sT{{\mathbb{T}}}

\newcommand{\R}{\mathbb{R}}

\DeclareMathOperator*{\argmax}{arg\,max}

%% file: contributions.tex
\definecolor{color0}{HTML}{FFFFFF}
\definecolor{color1}{HTML}{FFEDE5}
\definecolor{color2}{HTML}{FEE5D8}
\definecolor{color3}{HTML}{FED9C9}
\definecolor{color4}{HTML}{FDCAB5}
\definecolor{color5}{HTML}{FCBBA1}

\newcommand*{\mybox}[2]{{\setlength{\fboxsep}{0.15pt}\colorbox{#1}{\strut #2}}}

\begin{figure*}
\centering
\begin{tabular}{@{}p{13.5cm}@{\hspace{0.2cm}}p{13.5cm}@{}}
%  \resizebox{0.49\textwidth}{!}{%
  \small 
 \textbf{Hallucination}\\
 \midrule
 \small Sentence 1 \\ 
 \small
 \mybox{color0}{\strut{$\_$$\_$eng$\_$Latn$\_$$\_$}} \mybox{color0}{\strut{I}} \mybox{color0}{\strut{love}} \mybox{color0}{\strut{being}} \mybox{color4}{\strut{chubby.}} \mybox{color0}{\strut{$<$/s$>$}}  \mybox{color0}{\strut{$<$/s$>$}} \mybox{color1}{\strut{$\_$$\_$fra$\_$Latn$\_$$\_$}} \mybox{color0}{\strut{J'adore}} \mybox{color0}{\strut{être}} \mybox{color2}{\strut{une}} \mybox{color5}{\strut{grosse}} \mybox{color0}{\strut{\bf{chatte}.}}\\ %part8 unfiltered 9564
  \midrule
  \small 
 \textbf{Mistranslation}\\
 \midrule
 \small Sentence 2 \\
 \midrule
 \small
 \mybox{color0}{\strut{$\_$$\_$eng$\_$Latn$\_$$\_$}} \mybox{color0}{\strut{Hi!}} \mybox{color0}{\strut{I'm}} \mybox{color0}{\strut{a}} \mybox{color5}{\strut{gangly}} \mybox{color1}{\strut{child.}} \mybox{color0}{\strut{$<$/s$>$}} \mybox{color0}{\strut{$<$/s$>$}} \mybox{color4}{\strut{$\_$$\_$cat$\_$Latn$\_$$\_$}} \mybox{color0}{\strut{Hola,}} \mybox{color0}{\strut{sóc}} \mybox{color0}{\strut{un}} \mybox{color4}{\strut{nen}} \mybox{color0}{\strut{\bf{malparit}.}}\\
  \midrule
  \small
%\textbf{Semantically related translation, wrong level of offensiveness}\\
%\midrule
% \small Sentence 3 \\
% \midrule
% \small
% \mybox{color0}{\strut{$\_$$\_$eng$\_$Latn$\_$$\_$}} \mybox{color0}{\strut{Hi}} \mybox{color0}{\strut{I'm}} \mybox{color0}{\strut{a}} \mybox{color5}{\strut{queer}} \mybox{color2}{\strut{sibling.}} \mybox{color0}{\strut{$<$/s$>$}} \mybox{color0}{\strut{$<$/s$>$}} \mybox{color1}{\strut{$\_$$\_$spa$\_$Latn$\_$$\_$}} \mybox{color0}{\strut{Hola}} \mybox{color0}{\strut{soy}} \mybox{color0}{\strut{un}} \mybox{color2}{\strut{hermano}} \mybox{color0}{\strut{\bf{marica}.}}\\
%\mybox{color0}{\strut{$\_$$\_$eng$\_$Latn$\_$$\_$}} %\mybox{color0}{\strut{I'm}} \mybox{color0}{\strut{not}} %\mybox{color0}{\strut{a}} \mybox{color5}{\strut{queer}} %\mybox{color1}{\strut{guy.}} \mybox{color0}{\strut{$<$/s$>$}} %\mybox{color0}{\strut{$<$/s$>$}} %\mybox{color3}{\strut{$\_$$\_$cat$\_$Latn$\_$$\_$}} %\mybox{color0}{\strut{No}} \mybox{color0}{\strut{sóc}} %\mybox{color0}{\strut{un}} \mybox{color2}{\strut{tipus}} %\mybox{color0}{\strut{\bf{maricona}.}}\\
% \midrule
% \small
 %\textbf{Semantically related translation, correct level of offensiveness}\\
 \textbf{Correct translation}\\
\midrule
 \small Sentence 3 \\
 \midrule
 \small
%  \mybox{color0}{\strut{$\_$$\_$eng$\_$Latn$\_$$\_$}} \mybox{color0}{\strut{Hi!}} \mybox{color0}{\strut{I'm}} \mybox{color0}{\strut{a}} \mybox{color0}{\strut{person}} \mybox{color0}{\strut{with}} \mybox{color5}{\strut{dyslexia.}} \mybox{color0}{\strut{</s>}} \mybox{color0}{\strut{</s>}} \mybox{color2}{\strut{$\_$$\_$cat$\_$Latn$\_$$\_$}} \mybox{color0}{\strut{Hola!}} \mybox{color0}{\strut{Sóc}} \mybox{color0}{\strut{una}} \mybox{color0}{\strut{persona}} \mybox{color3}{\strut{amb}} \mybox{color0}{\strut{dislèxia.}}\\
   \mybox{color0}{\strut{$\_$$\_$eng$\_$Latn$\_$$\_$}} \mybox{color0}{\strut{Hi!}} \mybox{color1}{\strut{I'm}} \mybox{color0}{\strut{a}} \mybox{color5}{\strut{50-year-old}} \mybox{color1}{\strut{woman.}} \mybox{color0}{\strut{$<$/s$>$}} \mybox{color0}{\strut{$<$/s$>$}} \mybox{color3}{\strut{$\_$$\_$cat$\_$Latn$\_$$\_$}} \mybox{color0}{\strut{Hola,}} \mybox{color0}{\strut{sóc}} \mybox{color0}{\strut{una}} \mybox{color3}{\strut{dona}} \mybox{color0}{\strut{de}} \mybox{color0}{\strut{\bf{50}}} \mybox{color0}{\strut{\bf{anys}.}}\\
\midrule

  \end{tabular}
  \caption{Examples of translations in English-to-French, English-to-Spanish or English-to-Catalan.
  %Each sentence shows the input attribution to the word in bold  
  Sentences show input attributions for bold words in the cases of hallucination (sentence 1); mistranslation (sentence 2); %a semantically related translation but wrong level of offensiveness (sentence 3); 
  and a correct translation (sentence 3).  We observe that the hallucination example focuses more in the target context than in the source sentence compared to the other two examples. \label{fig:hallucinationmistranslation}}
\end{figure*}